\documentclass[final]{cvpr}

\usepackage{times}
\usepackage{epsfig}
\usepackage{graphicx}
\usepackage{amsmath}
\usepackage{amssymb}
\usepackage{graphicx}
\usepackage{multirow, makecell, caption}

\usepackage[pagebackref=true,breaklinks=true,colorlinks,bookmarks=false]{hyperref}

\def\cvprPaperID{2099} % *** Enter the CVPR Paper ID here
\def\confYear{CVPR 2021}

\begin{document}

%%%%%%%%% TITLE
\title{Kinematics-Guided Reinforcement Learning for
\\Object-Aware 3D Ego-Pose Estimation}

\author{
Zhengyi Luo \thanks{indicates equal contribution.} \textsuperscript{\rm 1}
\and
Ryo Hachiuma \footnotemark[1] \textsuperscript{\rm 2} 
\and
Ye Yuan \textsuperscript{\rm 1} 
\and
Shun Iwase \textsuperscript{\rm 1} 
\and
Kris M. Kitani \textsuperscript{\rm 1} 
\and

\textsuperscript{\rm 1} Carnegie Mellon University 
\textsuperscript{\rm 2} Keio University \\
{\tt\small zluo2@cs.cmu.edu, ryo-hachiuma@hvrl.ics.keio.ac.jp, \{yyuan2, siwase,kkitani\}@cs.cmu.edu}

}

% \and

\maketitle

\begin{abstract}
We propose a method for incorporating object interaction and human body dynamics into the task of 3D ego-pose estimation using a head-mounted camera. 
We use a kinematics model of the human body to represent the entire range of human motion, and a dynamics model of the body to interact with objects inside a physics simulator. 
By bringing together object modeling, kinematics modeling, and dynamics modeling in a reinforcement learning (RL) framework, we enable object-aware 3D ego-pose estimation. 
We devise several representational innovations through the design of the state and action space  to incorporate 3D scene context and improve pose estimation quality. We also construct a fine-tuning step to correct the drift and refine the estimated human-object interaction. This is the first work to estimate a physically valid 3D full body interaction sequence with objects (\emph{e.g.}, chairs, boxes, obstacles) from egocentric videos. Experiments with both controlled and in-the-wild settings show that our method can successfully extract an object-conditioned 3D ego-pose sequence that is consistent with the laws of physics.
\end{abstract}

\section{Introduction}

\begin{figure}[t]
    \begin{center}
    \includegraphics[width=\linewidth]{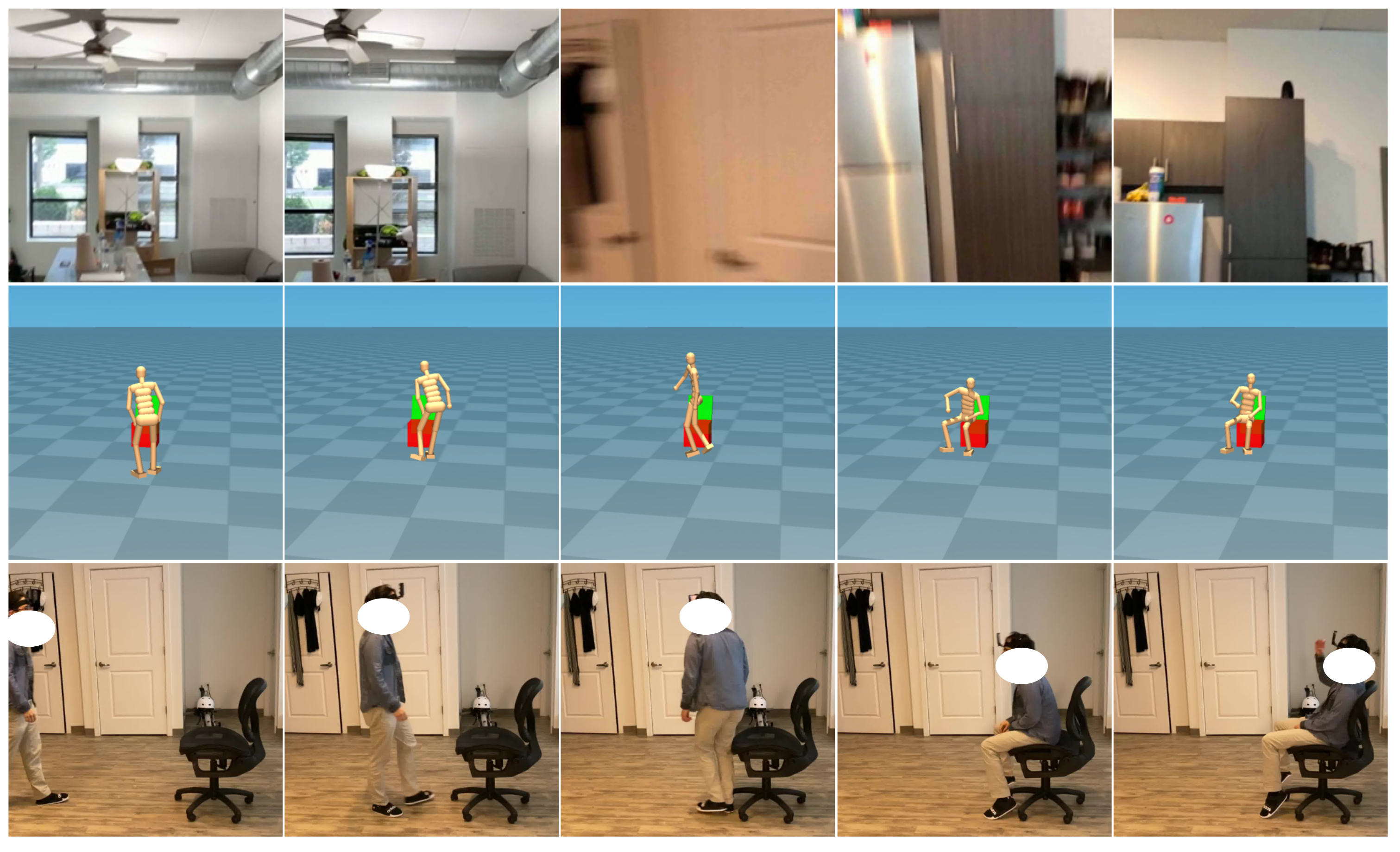}
    \end{center}
    \caption{Given an in-the-wild egocentric video, our method can infer physically valid 3D human pose and human-object interaction.  \textbf{Top}: input egocentric video. \textbf{Middle}: estimated 3D human pose and human-object interaction. \textbf{Bottom}: reference third person view.}
    \label{fig:teaser}
\end{figure}

From a video captured by a single head-mounted wearable camera (\emph{e.g.}, smartglasses, action camera, body camera), we want to infer the wearer's 3D pose and interaction with objects in the scene, as shown in Figure \ref{fig:teaser}. This is crucial for applications such as virtual/augmented reality, sports analysis, medical monitoring, etc., where third-person views are often unavailable and high-quality estimates of complex and dynamic human motion are needed. However, this task is challenging since the wearer's body is often unseen from a first-person view and the body motion needs to be inferred solely based on the visual context captured by the front-facing video. Furthermore, modeling physically realistic human-object interactions requires not only estimating the kinematic motion of the wearer, but also modelling the physical dynamics--the objects need to react to the forces applied by the human action in a physically realistic way. In this paper, we show that it is possible to infer accurate human motion and human-object interaction from a single, forward-facing wearable camera.

First and foremost, we address the important issue of using either 1) kinematics or 2) dynamics-based human pose estimation approaches. 1) Kinematics-based approaches study motion without regard to the forces that cause it. These methods directly output the joint angles of the human model and ignore physics (\emph{e.g.,} how much force is needed at each joint to hold the pose). They can often achieve better pose estimates but produce results that may violate physical constraints (\emph{e.g.,} joints bending the wrong way). Moreover, a kinematics-based model can not faithfully emulate human-object interactions--no physically realistic grasping, pushing, stepping, etc. can be performed without simulating physics. 2) Dynamics-based approaches represent methods that study motions that result from forces. They use a physics simulator and output joint torques to control the humanoid inside the simulator. Thus, these approaches output physically realistic human poses and can lead to convincing human-object interaction (pushing an object will make it move accordingly). However, since no ground truth joint torques are available, dynamics-based approaches are hard to learn and generalize. In this work, we argue that a hybrid approach is needed and propose using the output of the kinematics model as an additional signal to aid the training process and improve the performance of the dynamics model. Instead of training a dynamics model to directly map from the visual context to target joint angles like in prior works \cite{Yuan_2018_ECCV, yuan2019ego}, we propose to train an additional kinematics pose estimator and employ a novel action representation where the RL policy network is tasked to compute the residual pose against the output from a trained kinematics model. Such a formulation makes use of the accurate pose estimation from a kinematics model while using a dynamics model to refine the estimated pose to obey the laws of physics.

The second key missing piece is the scene context. Prior works \cite{Yuan_2018_ECCV, yuan2019ego, Jiang2016SeeingIP} have long omitted the semantic scene context from the first-person view due to the complexity of modeling physically correct human-object interactions. However, the presence of objects in the first-person view can often provide a strong prior over the expected human behavior. For example, combining the presence of a chair and the motion of moving forward, turning around, and bending down, we can strongly infer the action of sitting down. Thus, it is imperative for learning models to draw upon this contextual information to make an educated guess about human motion and human-object interaction. To this end, we incorporate the 6-degree-of freedom (DoF) pose of the main object of interest into the state representation of our RL model to make our model aware of the objects' states in the scene.  

% , and for the first time, bring in objects in the scene to enable object-aware 3D ego-pose estimation. We propose key innovations in designing the state \& action space of our RL model and a three-stage process to estimate human pose from egocentric videos. 

Another obstacle is the dynamics mismatch between the real world and simulation. As noted by prior work \cite{yuan2019ego}, global position and orientation drifts can often be observed over a long horizon of simulation. Such drift can lead to a catastrophic failure of human-object interaction: the humanoid can completely miss the box to push or the chair to sit on. To make sure the correct human-object interaction can be simulated, we propose a fine-tuning step against video evidence to correct the drift in the root trajectories. We use a monocular camera tracking technique, such as Visual Inertial Odometry (VIO) \cite{Wang2017Deep,Engel2013Semi}, to extract the camera motion and fine-tune our learned policy to match against it. The fine-tuning step alleviates the mismatch between the estimated and real-world global trajectories, leading to a successful human-object interaction. 

% Furthermore, as the reward with only the head information may cause an unrealistic pose, we introduce two regularization terms: the kinematic regularization and the action regularization. The kinematic regularization enforces similarity between the generated pose sequence and the kinematic pose. The action regularization makes the action estimated from the policy network to be close to it estimated from the pretrained policy.

% We made a dataset - we should mention but not really a contribution
As there is no public available egocentric video dataset that contains the 3D full-body pose and the 3D object pose, we capture a large-scale motion capture (MoCap) dataset in which the person wears a head-mounted camera and interacts with various objects. We capture three types of interactions: 1)sitting on (and standing up from) a chair, 2)pushing a box, and 3)avoiding obstacles while walking. We also capture an in-the-wild dataset that contains the same set of interactions.

% summary of contributions - important
In summary, we tackle the challenging task of extrapolating 3D human motion and human-object interaction from egocentric videos. Our contributions are as follows: (1) We are the first to propose a DeepRL based method for physically valid 3D pose and human-object interaction estimation from egocentric videos. (2) We propose to use a hybrid of kinematics and dynamics approaches and employ a novel action representation in which the dynamics-based model outputs the residual of the action against a learned kinematics-based model. (3) We propose a fine-tuning step to reduce the drift and refine our estimation based on captured video evidence. (4) We experiment with a self-made large-scale MoCap dataset \& an in-the-wild dataset, and show that our model outperforms other state-of-the-art methods on several pose-based and physics-based metrics, while generalizing to the in-the-wild settings. Upon visual inspection, our framework can not only recover the 3D human motion from egocentric videos, but also simulate physically correct human-object interactions. 

% Ethical considerations???

\begin{figure*}[t]
\begin{center}
\includegraphics[width=0.7\hsize]{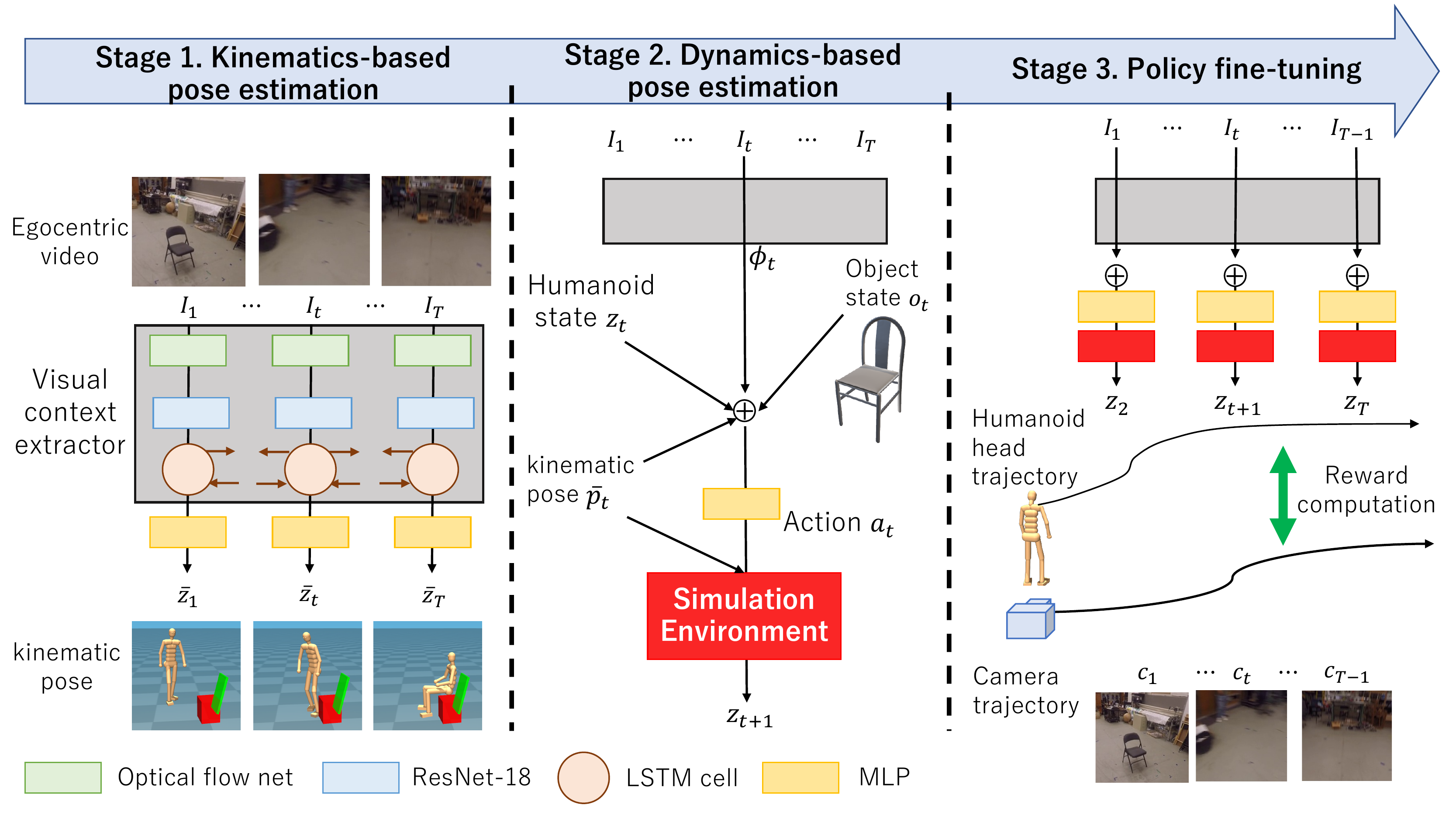}
\end{center}
\caption{The overview of our proposed pipeline. At first, a pose regressor will estimate the kinematic pose from video evidence. Then an object-aware dynamics-based model will perform and correct the estimated kinematic pose in a physics simulator. Finally, a fine-tuning step corrects the root drifts and produces the final physically valid pose and human-object interaction. }
\label{fig:overview}
\end{figure*}

\section{Related Works}
This section is divided into two parts. First, we will discuss kinematics-based approaches for 3D human pose estimation from egocentric videos. Second, we will discuss recent advancements in dynamics-based humanoid control methods. 

\subsection{3D human pose estimation from egocentric videos}
The task of estimating the 3D human pose from third-person videos is well researched in the computer vision community \cite{Rogez2019LCR,Pavllo20193D,Habibie2019In,Moon2019MPPE, Kolotouros2019LearningTR, Kocabas2020VIBEVI, Luo20203DHM}. These methods are also all kinematics-based and often result in physically invalid motions. 

On the other hand, there are only a handful of attempts at estimating 3D full body poses from egocentric videos, due to the ill-posed nature of this task. Most existing methods still assume partial visibility of body parts in the image \cite{tome2019xr,Rhodin2016EgoCap,xu2019mo2cap2}, often through a downward-facing camera. Among works where the human body is mostly not observable,  \cite{Jiang2016SeeingIP,Yuan_2018_ECCV,yuan2019ego,Ng2019You}, \cite{Jiang2016SeeingIP} uses a kinematics-based approach where they construct a motion graph from the training motions and recover the pose sequence by solving the optimal pose path. \cite{Ng2019You} focuses on modeling person-to-person interactions from egocentric videos and infers the wearer's kinematic pose conditioning on the other person's pose. \cite{Yuan_2018_ECCV, yuan2019ego,isogawa2020optical}, on the other hand, use dynamics-based approaches where a RL-based agent is tasked to perform physically valid human motions. In comparison, our work combines kinematics-based and dynamics-based approaches to achieve both accurate and physically valid pose estimation. In contrast to previous works, we also model human-object interactions such as sitting on a chair and pushing a box on the table. To the best of our knowledge, we are the first approach to estimate the 3D human poses from egocentric video while factoring in human-object interactions.

\subsection{Humanoid control for object manipulation}
Our work is also connected to controlling humanoids to interact with objects in a physics simulator \cite{Peng2018DeepMimic,Peng2018SFV, chao2019learning,merel2019reusable,yuan2020residual}. The core motivation of this line of work is to learn the necessary dynamics to imitate realistic human motion in a physics simulation. 

\cite{chao2019learning} proposes a hierarchical reinforcement learning approach to generate realistic sitting motion. The authors manually decompose high-level actions (such as sitting) to low-level actions (such as walking, turning, and sitting) from which the proposed meta-controller can stochastically select. They do not aim to estimate the full body pose from egocentric videos, and their motion is limited to sitting. Our approach can not only replicate a diverse set of sitting motions, but also ground sitting motion on video evidence. \cite{merel2019reusable} proposes an approach for enabling humanoid full-body manipulation and locomotion in simulation. They use a {\it phased task} in which the task policy is trained to solve different stages of the task and show impressive results in human-object manipulation. Similarly, their generated motion is not grounded on video evidence. \cite{Soohwan2019learning} uses an action representation in which the target pose is the sum of the kinematic pose and the output of the policy network. Inspired by their work, we also employ this residual action representation to accelerate training and improve stability. Our task is also related to DeepMimic \cite{Peng2018DeepMimic} and its video variant \cite{Peng2018SFV}. DeepMimic has shown remarkable results in imitating human locomotion skills and is able to combine learned skills to achieve different tasks. However, human-object interaction and video grounding are not considered in their approach.

\section{Method}

The problem of 3D body pose estimation from egocentric videos can be formulated as follows: from a wearable camera footage $I_{1:T}$, we want to estimate the person's pose sequence $p_{1:T}$. We propose a three-step method: 1) A kinematic pose pose regression,  2) object-aware pose correction using dynamics, 3) a fine-tuning step. The overview of our proposed method is depicted in Figure \ref{fig:overview}. 

% First, we train a regression model which predicts the kinematic pose ${\bar p}_{1:T}$ from a egocentric video sequence $I_{1:T}$. This model does not consider the laws of physics like causal forces or actuation constraints, and the network is easier to train compared to the dynamics-based models. Secondly, we train the RL dynamics-based model that produces physically-valid poses $p_{1:T}$. In this stage, the task is formulated as an Markov decision process(MDP) that operates inside a physics simulation. The goal is to learn a optimal control policy that takes in the state of the current environment and output the next action. The physics simulation will take in the action and provides the state for the next time-step. Finally, a fine-tuning step is performed to correct the accumulated error in global position and orientation by the previous step. The key idea is that from the captured camera sequence, the $6$DoF camera trajectory can be obtained by using Visual Inertial Odometry and be used as pesudo-ground truth head pose. In the follow sections, we will dive into each individual component in detail.

\subsection{Pose Estimation using Kinematics}
To recover the kinematic 3D human pose from egocentric videos, we train a regressor that predicts the pose ${\bar p}_{1:T}$ from the input video sequence $I_{1:T}$. Specifically, at first, a long-short-term-memory (LSTM) based visual context extractor is used to extract visual information from egocentric videos: $I_{1:T} \rightarrow \phi_{1:T}$. Then a multilayer perceptron (MLP) is used to produce the kinematic states from the visual context $\phi_{1:T} \rightarrow \hat{z}_{1:T}$. These states ${\bar z}_t$ consist of the human pose (position on the horizontal plane, orientation of the root, and the joint angles) and velocities (linear and angular velocities of the root and joint velocities). From the kinematic state, we can recover the full kinematic pose sequence ${\bar p}_{1:T}$ that consists of root position, root orientation, and joint angles. We employ mean squared error (MSE) as the loss function to train the regressor: $L(\xi)=\frac{1}{T}\sum_{t=1}^T\|{\mathcal F}(I_{1:T})_t-{\hat z}_t\|^2$, where $\xi$ are the parameters of $\mathcal F$ and ${\hat z}_t$ is the ground truth kinematic state from MoCap. In general, this model does not consider the laws of physics like forces or actuation constraints, so the network is easier to train compared to dynamics-based models.

\subsection{Object-Aware Pose Estimation using Dynamics}
We formulate the task of estimating a physically valid pose sequence $p_{1:T}$ from egocentric RGB images $I_{1:T}$ as a Markov Decision Process (MDP) defined as a tuple ${\mathcal M}=\langle S, A, P, R, \gamma\rangle$ of states, actions, transition dynamics, reward function, and discount factor. The state $S$ and the transition dynamics $P$ are provided by the physics simulator, and the action $A$ and the reward $R$ are computed by the policy $\pi$. At each time step $t$, the agent in state $s_t$ takes an action sampled from the policy $\pi(a_t|s_t)$ while the environment generates the next state $s_{t+1}$ based on that action through physics simulation. Comparing the resulting state of the humanoid with the ground truth, the agent will receive a reward $r_t$. This process repeats until some termination condition is triggered, such as when the time horizon is reached or the humanoid falls onto the ground. We employ Proximal Policy Optimization (PPO) \cite{schulman2017proximal} to calculate the optimal policy $\pi^\ast$ that maximizes the expected discounted return $E[\sum_{t=1}^T \gamma^{t-1}r_t]$. At the test time, we roll out the policy $\pi^{\ast}$ to generate a state sequence $s_{1:T}$ from which we extract the output pose sequence $p_{1:T}$. 

To enable object-aware motion estimation from egocentric videos that abide by the laws of physics, we innovate on two key points upon prior dynamics-based models: (1) we factor in the object pose into our RL agent's state representation, (2) we use the result from the pre-trained regressor from the previous step as an additional input to the RL model. Each of the MDP elements is defined as follows:

\subsubsection{State.} The state $s_t$ at time step $t$ consists of the humanoid state $z_t$, the visual context $\phi_t$, the kinematic pose state ${\bar q}_t$, and the object state $o_t$: $s_t = <z_t, \phi_t, o_t, {\bar q}_t>$. $z_t$ consists of the humanoid pose $q_t$ (position and orientation of the root + joint angles) and velocity $v_t$ (linear and angular velocities of the root  + joint velocity). ${\bar q}_t$ is the output of the kinematic pose regressor $\mathcal{F}$. Here, we factor in the 6DoF object pose $o_t$ as an additional input to the control policy to enable object-aware 3D pose estimation. 

\subsubsection{Action.} The action $a_t$ specifies the target joint angles for the proportional-derivative (PD) controller  controller \cite{Tan2011Stable} at each degree of freedom (DoF) of the humanoid joints except for the root (Hip). We use a novel residual action representation:
\begin{equation}
q^d_t = {\bar q}_t + \Delta q^d_t,
\end{equation}
 $q^d_t$ is the final PD target, $\Delta q^d_t$ is the output (action $a_t$) of the control policy $\pi$, and ${\bar q}_t$ denotes the predicted pose of the kinematic pose regressor $\mathcal F$. For joint $i$, the torque to be applied is computed as $\kappa^i = k_p^i(q^d_t-p_t^i)-k_d^i v_t^i$ where $k_p$ and $k_d$ are manually specified gains. Compared to directly estimating the target joint angle \cite{yuan2019ego}, predicting the residual of the target pose against the kinematic pose ${\bar q}_t$ offers a better starting point for the RL policy and results in faster convergence and improved stability.

\subsubsection{Policy.} The policy $\pi_\theta(a_t|s_t) = \pi_\theta(a_t|z_t, \phi_t, o_t, {\bar q}_t)$ is represented by a Gaussian distribution with a fixed diagonal covariance matrix $\Sigma$. We use a MLP parametrized by $\theta$ as our policy network to map the state $s_t$ to the predicted mean $\mu_t$. 

\subsubsection{Reward function.} The reward function is as follows:
\begin{equation}
r_t = w_p r_p + w_e r_e + w_{rv} r_{rv} + w_{rq} r_{rq} + w_{rp} r_{rp},
\end{equation}
where $w_p, w_e, w_{rv}, w_{rq}, w_{rp}$ are the weights of each reward. Our reward is similar to DeepMimic \cite{Peng2018DeepMimic}, with the exception that we separate the root reward from the pose reward to better motivate the model to match the ground truth root trajectory. More importantly, unlike the conventional object manipulation control methods \cite{Peng2018DeepMimic, merel2019reusable, Peng2018SFV}, we do not set any manually designed goal reward for each specific task and only use the pose reward to match ground truth poses to handle multiple interactions. The pose reward $r_p$ measures the difference between the generated pose $q_t$ and the ground truth pose $\hat{q}_t$ in quaternion for each joint on the humanoid except for nonroot joints. The end-effector reward $r_e$ computes the distance between the estimated end-effector (foot, hand, head) position $e_t$ and the ground truth position $\hat{e}_t$. The root velocity reward $r_{rv}$ penalizes the deviation of the estimated root's linear $l_t $ and angular $\omega_t$ velocity from the ground truth $\hat{l}_t $ \& $\hat{\omega}_t$. The ground truth velocity is computed from the data via finite difference. The root position \& orientation rewards $r_{rp}$ \& $r_{rq}$ compute the difference between the generated 3D root position $p_t$ \& orientation $q_t$ and the ground truth $\hat{p}_t$ \& $\hat{q}_t$ in the world coordinate frame:

\begin{align}
r_p &= \exp\left [-5.0\left( \Sigma_j\|{\bar q}^j_t \ominus q^j_t \|^2\right)\right], \\
r_{e} &=\exp \left[-4.5\left(\Sigma_e\left\|e_{t}-\hat{e}_{t}\right\|^{2}\right)\right], \\   
r_{rv} &=\exp \left[-\left\|l_{t}-\hat{l}_{t}\right\|^{2}-0.1\left\|\omega_{t}^{r}-\hat{\omega}_{t}^{r}\right\|^{2}\right],\\
r_{rq} &=\exp \left[-40\left(\left\|q_{t}^{r} \ominus \hat{q}_{t}^{r}\right\|^{2}\right)\right],  \\  
r_{rp} &= \exp \left[-45\left(\left(p_{t}-\hat{p}_{t}\right)^{2}\right)\right] .   
\end{align}

\subsubsection{Initial state estimation.}  During training, we set the initial humanoid state $z_1$ and the object state $o_1$ to the ground truth ${\hat z}_1$, ${\hat o}_1$. At the test time, the starting states of the humanoid and objects are given by the kinematic pose regressor and an off-the-shelf 6DoF object pose estimator, respectively.

\subsection{Fine-tuning of the policy network}
\label{sec:ft}
As mentioned previously, at the test time, the output of the RL model will drift from the ground truth root position and orientation, causing the humanoid to trip and the human-object interaction to fail. Moreover, as the training data cannot cover all object state (position \& orientation), it is difficult for the policy to generalize to unseen states. 

To overcome this problem, we propose to fine-tune the policy $\pi_\theta$ with test-time video evidence. While capturing the egocentric video, the $6$DoF camera motion (position ${\hat h}^p_t$ and orientation ${\hat h}^q_t$) can be recovered using VIO techniques. Using the camera trajectory as an approximation to head motion, we can fine-tune the trained dynamics-based pose estimator to come up with physically valid pose estimates that conform to the extracted camera trajectory from the video. However, a naive fine-tuning step that only attempts to match humanoid motion with the tracked camera motion may lead to unnatural human poses since the policy may forget how to produce valid human poses. To overcome this issue, we introduce two types of regularization, 1) the fine-tuned pose estimate ${\bar p}_t$ must not deviate from the input kinematics result too drastically. 2) we regularize the action space by penalizing the difference of the output $\tilde\mu$ from the pretrained policy network $\tilde\pi$ and $\mu$ from the fine-tuned policy $\pi$. This way, our policy is forced to not deviate too much from its original behavior while trying to conform to the tracked camera trajectory. The overall reward at the fine-tuning stage is as follows:
\begin{equation}
\label{eq:ft_reward}
{\hat r}_t = w_{hp} r_{hp}+w_{hq}r_{hq} + w_{hv}r_{hv} + w_p \lambda_t r_p + w_a (1-\lambda_t)r_a
\end{equation}
where $r_{hp}, r_{hq}, r_{hv}, r_p', r_a$ are the rewards for head position, head orientation, head linear and angular velocity, pose, and action, respectively. $w_{hp}$, $w_{hq}$, $w_{hv}$, $w_p$, $w_a$ are the weighting factors. The head position $r_{hp}$, orientation $r_{hq}$, velocity $r_{hv}$, and rewards are similar to their root counterparts $r_{rp}$, $r_{rq}$, and $r_{rv}$. They are computed similarly and here we use the extracted camera motion as an approximation to ground-truth head trajectory. The pose reward $r_p'$ penalizes the difference of the generated pose $p_t$ and the kinematic pose ${\bar p}_t$ for joints except for the root joint.  The action reward penalizes the difference between the mean action of the pretrained policy $\tilde\mu$ and the mean action of the current policy $\mu$. Furthermore, we introduce an adaptive weighting factor $\lambda_t$ on the action reward: if the kinematic regressor $\mathcal F$ generates unrealistic poses, the policy should avoid imitating that pose. To this end, we use a confidence value that puts more weight to the kinematic reward if the model has high confidence on the estimated kinematic pose (imitates the estimated kinematic pose more) and leans toward the action reward if the kinematic pose is deemed unrealistic (sticks to the original action more). We calculate this value based on the head linear velocity in local  coordinates: where ${\hat h}^{lv}_t$ denotes the linear velocity of the camera and the ${\hat h}^{lv}_t$ denotes the linear velocity of the humanoid head in the local frame, respectively:

\begin{align}
r_{hp}  &= \exp\left [-10.0\left( \|{\hat h}^p_t - h^p_t \|^2\right)\right], \\
r_{hq} &= \exp\left [-10.0\left( \|{\hat h}^q_t \ominus h^q_t \|^2\right)\right],\\
r_{hv} &= \exp\left [-0.1\left(\|{\hat h}^v_t - h^v_t \|^2\right)\right], \\
r_{p}'\left(\bar{q}_{t}, q_{t}\right) &=\exp \left[-5.0\left(\Sigma_{j}\left\|\bar{q}_{t}^{j} \ominus q_{t}^{j}\right\|^{2}\right)\right], \\
r_a(\mu, \tilde\mu) &= \exp\left [-1.0\left( \|\tilde\mu - \mu \|^2\right)\right], \\
\lambda_t &= \exp\left [-0.1\left(\|{\hat h}^{lv}_t - h^{lv}_t \|^2\right)\right].
\end{align}
 
 \begin{figure*}[tb]
\begin{center}
\includegraphics[width=\textwidth]{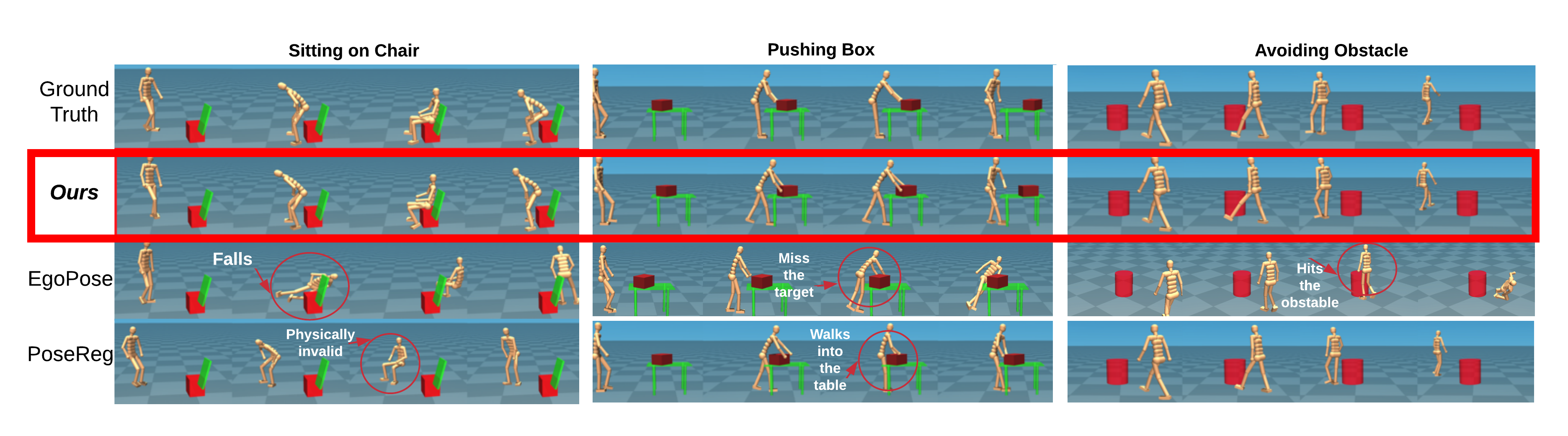}
\end{center}
\caption{Results of 3D pose and human-object interaction estimation from egocentric videos.}
\label{fig:qualitative}
\end{figure*}

\section{Experimental Setup}
\subsection{Dataset} 
\label{wilddata}

As there is no public dataset available containing the ground-truth full-body human pose and object pose annotations where the person interacts with objects, we record a large-scale egocentric video dataset inside a Mocap studio. It includes three subjects and each subject is asked to wear a head-mounted camera and performs various complex human-object interactions for multiple takes. The actions consist of sitting on (standing up from) a chair, avoiding obstacles, and pushing a box. There are four categories of objects: chairs, tables, boxes, and obstacles. MoCap markers are attached to the camera wearer and the objects to get the 3D full-body human pose and $6$DoF object pose. To diversify the ways actions are performed, we ask the actors to vary their performance for each action. We also sample the actors' initial facing directions and position uniformly from a circle of 3-meter radius. Each take is about six seconds long; $16$ to $24$ sequences exist for each combination of action and subject. As a result, our MoCap dataset consists of about $250$ sequences. We use an $80\mbox{--}20$ train test data split on this MoCap dataset. To further showcase the generalization of our method, we also collect an in-the-wild dataset where an additional subject is tasked to perform similar tasks in an everyday setting wearing a head-mounted iPhone. Here, we use Apple's ARKit to provide camera position \& orientation tracking as well as 6 DoF object pose estimation. The in-the-wild dataset has 15 takes in total, each lasting about 6 seconds. For in-the-wild, we use a third-person camera to capture a synchronized side-view of the person and use an off-the-shelf 3D human pose estimator \cite{Luo20203DHM} to estimate 3D poses as pseudo-ground truth for evaluation. 
% Details about both of the datasets can be found in the supplementary materials. 

\subsection{Evaluation metrics}
To quantitatively evaluate for 3D pose accuracy and its physical correctness, we employ the following metrics:

{\noindent \textbf{Root error (${\bf E}_{root}$)}}: a pose-based metric that measures the difference between the ground truth and the generated root pose, both represented as the $4 \times 4$ transformation matrix $M_t(q^r_t|g_t)$ composed of the root orientation $q^r$ and the translation $g$. The error is calculated using the Frobenious norm: $\frac{1}{T}\sum_{t=1}^T\|{ I} - (M_t{\hat M}_t^{-1})\|_F$ where $I$ is the identity matrix.

 \noindent \textbf{Pose Error (${\bf E}_{joint}$)}: a pose-based metric that measures the difference between the estimated pose sequence $p_{1:T}$ and the ground truth ${\hat y}_{1:T}$, both represented as Euler angles in radians:  $\frac{1}{T}\sum_{t=1}^T\|p_t-{\hat p}_t\|_2$. This metric does not consider the root pose and only compares the body joint angles. 

\noindent{\textbf{Mean Per Joint Position Error (${\bf E}_{mpjpe}$)}}: a pose-based metric used for our in-the-wild evaluation. As off-the-shelf pose estimator \cite{Luo20203DHM} from third-person view uses a different human model \cite{Loper2015SMPLAS} than ours, we cannot directly compare the joints' angles. Thus, we find the common joints (15 in total) between the two human models and directly compare their 3D \textbf{positions} in the global coordinate frame. Denote pseudo-ground truth joint position and the estimated joint positions as $J_t, \hat{J_t} \in R^{15 \times 3}$. The metric is calculated as $\frac{1}{T}\sum_{t=1}^T\|J_t-{\hat J}_t\|_2$, measured in millimeters. This metric factors in the root pose as well as the joint angles. 

\noindent{\textbf{Velocity Error (${\bf E}_{vel}$)}}: a physics-based metric that measures the difference between the estimated joint velocity $v_{1:T}$ and the ground truth ${\hat v}_{1:T}$: $\frac{1}{T}\sum_{t=1}^T\|v_t-{\hat v}_t\|_2.$ where $v_t$ and ${\hat v}_t$ are computed by the finite difference. When ground-truth joint angles are available, this metric is calculated as the angular velocity in radians; for in-the-wild evaluation, it is measured in linear velocity in millimeters/second. 
\noindent{\textbf{Average Acceleration (${\bf A}_{accel}$)}}: a physics-based metric that uses the average magnitude of joint acceleration to measure the smoothness of the predicted pose sequence, calculated as $\frac{1}{T}\sum_{t=1}^{T}\|{\dot v}_t\|_1$ where ${\dot v}_t$ denotes the joint acceleration. Similar to the velocity error, for experiments with ground-truth pose available, this metric is calculated in angular acceleration in radians; for in-the-wild evaluation, it is measured in linear acceleration in millimeters/second$^2$.

More details about our network and the physics simulator (Mujoco) are provided in the supplementary material. 

\subsection{Baseline methods}
To show the effectiveness of our framework, we compare against two baseline methods: (1) the previous state-of-the-art method in this task, \textbf{EgoPose} \cite{yuan2019ego}, a dynamics-based approach that produces physically realistic motion but does not factor in object states, and (2) the previous best kinematics-based approach, also proposed in \cite{yuan2019ego}, which we will call \textbf{PoseReg}.

\begin{table*}[t]
\centering
\caption{\textbf{\textit{Single-subject}} quantitative results for pose-based and physics-based metrics per action}
\resizebox{0.8\textwidth}{!}{%
\begin{tabular}{l|c|c|c|c||c|c|c|c||c|c|c|c|c|c|r}
\hline
\multicolumn{1}{c|}{\multirow{2}{*}{}} & \multicolumn{4}{c||}{Sitting} & \multicolumn{4}{c||}{Avoiding}  & \multicolumn{4}{c|}{Pushing} \\ \cline{2-13} 
\multicolumn{1}{c|}{} & $\textbf{E}_{root}\downarrow$ & $\textbf{E}_{joint}$ $\downarrow$ &  $\textbf{E}_{vel}$ $\downarrow$ & $\textbf{A}_{accel}$ $\downarrow$ & $\textbf{E}_{root}$ $\downarrow$ & $\textbf{E}_{joint}$ $\downarrow$ &  $\textbf{E}_{vel}$ $\downarrow$ & $\textbf{A}_{accel}$ $\downarrow$ & $\textbf{E}_{root}$ $\downarrow$ & $\textbf{E}_{joint}$ $\downarrow$ &  $\textbf{E}_{vel}$ $\downarrow$ & $\textbf{A}_{accel}$ $\downarrow$ \\ \hline
    PoseReg & 1.151 & {\bf 0.915} & 6.431 & 12.609  & 0.682 & {\bf 0.638} & 6.152 & 12.126 & 0.909 & 0.825 & 6.621 & 13.058       \\ \hline
    EgoPose & 1.444 &  1.3445 & 6.634 &10.181  & 1.293 & 1.061 & 7.573 &11.945& 1.362 & 1.219  & 6.704 & 9.459   \\ \hline
    \textbf{Ours} & {\bf 0.607}& 1.011 & {\bf 5.084}& {\bf 5.489}& {\bf 0.372} & 0.723 & {\bf 5.889} & {\bf 7.081} & {\bf 0.377}& {\bf 0.820}& {\bf 5.257}& {\bf 5.986} \\
    \hline
\end{tabular}}
\label{tab:quantitativeSingle}
\end{table*}

\begin{table*}[t]
\centering
\caption{\textbf{\textit{Cross-subject}} quantitative results for pose-based and physics-based metrics per action}
\resizebox{0.8\textwidth}{!}{%
\begin{tabular}{l|c|c|c|c||c|c|c|c||c|c|c|c|c|c|r}
\hline
\multicolumn{1}{c|}{\multirow{2}{*}{}} & \multicolumn{4}{c||}{Sitting} & \multicolumn{4}{c||}{Avoiding}  & \multicolumn{4}{c|}{Pushing} \\ \cline{2-13} 
\multicolumn{1}{c|}{} & $\textbf{E}_{root}\downarrow$ & $\textbf{E}_{joint}$ $\downarrow$ &  $\textbf{E}_{vel}$ $\downarrow$ & $\textbf{A}_{accel}$ $\downarrow$ & $\textbf{E}_{root}$ $\downarrow$ & $\textbf{E}_{joint}$ $\downarrow$ &  $\textbf{E}_{vel}$ $\downarrow$ & $\textbf{A}_{accel}$ $\downarrow$ & $\textbf{E}_{root}$ $\downarrow$ & $\textbf{E}_{joint}$ $\downarrow$ &  $\textbf{E}_{vel}$ $\downarrow$ & $\textbf{A}_{accel}$ $\downarrow$ \\ \hline
    PoseReg & 1.403 & {\bf 1.775} & 7.870 & 11.355 & 1.326 & {\bf 1.469} & 7.976 & 10.768  & 0.812 & 1.643 & 7.260 & 11.355       \\ \hline
    EgoPose & 1.722& 1.887 & 7.986 & 13.943& 1.514 & 1.749 & 9.685 & 13.942 &1.708& 1.934 & 7.870 & 11.042    \\ \hline
    \textbf{Ours} & {\bf 0.756} & 1.850 & {\bf 6.085} & {\bf 5.880} & {\bf 1.112} & 1.476 & {\bf 7.368} & {\bf 5.446} & {\bf 0.449} & {\bf 1.631} & {\bf 6.407} & {\bf 5.605}\\ 
    \hline
\end{tabular}}
\label{tab:quantitativeCross}
\end{table*}

\begin{table*}[!thb]
\caption{\textbf{\textit{In the wild \&Cross-subject }} Quantitative results, evaluated against pseudo-ground truth poses from a third person view. Notice that the unit (millimeters) in this table is different from the previous tables where ground-truth annotation is available.} \label{tab:wildtable}
\centering
\resizebox{0.8\textwidth}{!}{%
\begin{tabular}{l|c|c|c||c|c|c||c|c|c|c|c|c|r}
\hline
\multicolumn{1}{c|}{\multirow{2}{*}{}} & \multicolumn{3}{c||}{Sitting} & \multicolumn{3}{c||}{Pushing}  & \multicolumn{3}{c|}{Avoiding} \\ \cline{2-10} 
\multicolumn{1}{c|}{} & $\textbf{E}_{mpjpe}$ $\downarrow$ &  $\textbf{E}_{vel}$ $\downarrow$ & $\textbf{A}_{accel}$ $\downarrow$ & $\textbf{E}_{mpjpe}$ $\downarrow$ &  $\textbf{E}_{vel}$ $\downarrow$ & $\textbf{E}_{accel}$ $\downarrow$ & $\textbf{E}_{mpjpe}$ $\downarrow$ &  $\textbf{E}_{vel}$ $\downarrow$ & $\textbf{A}_{accel}$ $\downarrow$ \\ \hline
    PoseReg   & 453.40 & 26.91  & 26.48 & 576.41 &23.29 & 13.55 & 1680.76 &35.49  & 13.28    \\ \hline
    Egopose   & 551.91 & 28.5 & 20.47 & 518.19 &26.48 & 22.73 &  1540.17 & 38.25 & 21.02      \\ \hline
    Ours   & \textbf{313.76}  & \textbf{18.50} & \textbf{4.79} &  \textbf{248.85} & \textbf{16.65} & \textbf{4.17} & \textbf{440.08} & \textbf{23.12} & \textbf{5.69}     \\ \hline
\end{tabular}}
\end{table*}

\section{Results}

\subsection{Subject-Specific Evaluation.}
In this evaluation, we train an individual model for each subject and evaluate on the test split of our MoCap dataset. From the quantitative results in Figure \ref{fig:qualitative}, we can see that PoseReg can produce accurate pose estimation, but does not obey the laws of physics (such as sitting in mid-air, walking into the table). EgoPose, on the other hand, often fails to perform the action correctly (falls down or hits the obstacle). Overall, our method (second row) produces 3D human poses closer to the ground truth (top-row) than any other baselines, and successfully performs human-object interaction. \textbf{To better visualize the quantitative results, please refer to our supplementary video. }

Table \ref{tab:quantitativeSingle} shows the quantitative comparison of our method with the two baseline methods. All results are average across all three subjects. For all actions, we observe that our method outperforms the two baselines across almost all actions and metrics, with occasional worse performance in joint angle estimation with the kinematics-based method (PoseReg). This is expected as PoseReg disregards physics and can estimate poses without constraints, while our method requires estimating a physically valid pose. We find that the humanoid controlled by EgoPose often falls down to the ground and collides with the object, resulting in high pose error. This is expected as EgoPose does not factor in object states into its pose estimation and suffers from error accumulation in global root positions. On the other hand, our object-aware state representation and our fine-tuning step ensure a correct human-object interaction, preventing falls and drifts. We can also observe the low velocity error and acceleration compared to the baseline method, which indicates that our residual action representation produces a more stable and smoother pose estimation. 

\subsection{Cross-Subject Evaluation.}
To further test the robustness, we perform cross-subject experiments where we train our model on two subjects and test on the remaining subject. This is a challenging setting, since people have unique styles and speeds for the same types of interactions. The quantitative results are summarized in Table \ref{tab:quantitativeCross}: our method again outperforms other baseline methods in almost all metrics. Especially for the smoothness of the pose (${\bf A}_{accel}$), our method estimates much smoother (2.0x) pose sequences than those generated from other baseline methods.

\subsection{In-the-wild Cross-subject Evaluation.} To demonstrate the generalization of our method to real-world use cases, we further test our method on an in-the-wild dataset. Since there is no ground-truth 3D poses available, we evaluate against pseudo-ground truth 3D pose extracted from a third-person view camera. As shown in Table \ref{tab:wildtable}, our method outperforms the baseline methods by a large margin, especially on avoiding and pushing actions where root drifts are prominent. Notice that this dataset is captured in a real-world setting using a different camera (iPhone) than used in the MoCap studio (GoPro), and our method is able to generalize and infer valid human poses and human-object interactions. 

\subsection{Quantitative Evaluation}
As motion is best seen in video, we refer readers to our \textbf{supplementary video} for a comprehensive quantitative analysis. 

\begin{table}[tb]
\caption{Ablation study for the action representation, the fine-tuning step, and fine-tuning reward design.}
\raggedleft
\resizebox{0.5\textwidth}{!}{%
\begin{tabular}{c}
\begin{tabular}{lrrrr}
\hline 
Action representation& ${\bf E}_{root} \downarrow$   & ${\bf E}_{joint} \downarrow$  & ${\bf E}_{vel} \downarrow$   & ${\bf A}_{accel} \downarrow$   \\ \hline
(a) PD target  & 1.753 & 2.049 & 7.912 & 12.502 \\
(b) Kinematic residual & {\bf 1.210} & {\bf 1.064} & {\bf 6.051} & {\bf 8.529} \\
\hline 
\end{tabular}
\\ \\
\begin{tabular}{lrrrr}
\hline 
Reward type  & ${\bf E}_{root} \downarrow$   & ${\bf E}_{joint} \downarrow$  & ${\bf E}_{vel} \downarrow$   & ${\bf A}_{accel} \downarrow$   \\ \hline
(b) No fine-tuning & 1.210 &  1.064 &  6.051 &  8.529 \\
(c) head& 0.366 & 0.920& 5.227& 6.477 \\
(d) head, kinematic regularization & 0.374 & 0.810 & 5.090 & {\bf 6.291} \\
(e) head, action regularization. & 0.361& 0.882 & 5.136& 6.436 \\ 
(f) head, action + kinematic regularization.& 0.340& 0.805 & 5.074& 6.351 \\ 
(g) Full reward (Ours) & {\bf 0.322} & {\bf 0.788} & {\bf 5.063} & 6.380 \\ 
\hline 
\end{tabular}
\end{tabular}
}
\label{tab:ablation}
\end{table}

\subsection{Ablation Study.}
To evaluate the importance of (1) our novel action representation and (2) our fine-tuning step and its reward design, we conduct an ablation study to quantify their benefits. This study is conducted for the first subject's sitting action. The results of the ablation study are presented in Table \ref{tab:ablation}. To investigate the importance of our action representation (b), we compare the representation employed by  \cite{yuan2019ego} (a) in which the policy outputs the target joint of PD control directly (instead of a residual of the kinematics \textbf{}model). The prediction accuracy is improved greatly in all metrics (31\%, 48\%, 23\%, and 31\%, respectively) after using our residual action representation.

To investigate the importance of the fine-tuning step and each of its reward terms, we train the comparison models with four types of rewards, (c) head (position, orientation, and velocity) reward, (d) head and kinematic regularization reward, (e) head and action regularization reward, (f) head, and kinematic and action regularization reward. Model (g) has the additional adaptive weighting factor $\lambda_t$ in addition to the reward (f). From Table \ref{tab:ablation}, without the fine-tuning step, the model performs the worst (b). Our reward (g) outperforms all partial rewards (c-f) in the metrics except for the acceleration metric. Combining each reward term results in the best performance model across all spectrum, showcasing the benefit of our fine-tuning step and its respective rewards. 

\subsection{Failure cases and limitations}
Although our method can produce realistic human pose and human-object interaction estimation from egocentric videos, we are still at the early stage of this challenging task. Our method performs well in the MoCap studio setting and out-performs state-of-the-art methods in-the-wild, but can still fail to produce natural pose estimation for in-the-wild videos. Due to the severe domain shift, the result of the pose regressor is poor on in-the-wild data and can lead to unnatural motion such as severe foot sliding. As a result, our fine-tuning stage is forced to produce unnatural poses (e.g, long steps, awkward turns) to prevent falling. Furthermore, our method is still limited to a predefined set of interactions where we have data to learn from. To enable pose and human-object interaction estimation for arbitrary actions, much further investigation is needed. 

\section{Conclusion}
In this paper, we tackle, for the first time, physically valid 3D pose estimation from an egocentric video while the person is interacting with objects. We collect a large-scale motion capture dataset to develop and evaluate our method, and extensive experiments have shown that our method outperforms all prior methods. Real-world experiments using an in-the-wild dataset further show that our method generalizes well to real-world use cases and can infer physically valid 3D human motion and human-object interaction from a front-facing camera feed.

{\small
\bibliographystyle{ieee_fullname}
\bibliography{egbib}
}

\end{document}